\documentclass[sn-mathphys,Numbered]{sn-jnl}

\usepackage{graphicx}%
\usepackage{multirow}%
\usepackage{amsmath,amssymb,amsfonts}%
\usepackage{amsthm}%
\usepackage{mathrsfs}%
\usepackage[title]{appendix}%
\usepackage{textcomp}%
\usepackage{manyfoot}%
\usepackage{booktabs}%
\usepackage{algorithm}%
\usepackage{algorithmicx}%
\usepackage{algpseudocode}%
\usepackage{listings}%
\usepackage{color}
\usepackage{subcaption}
\usepackage{booktabs}
\usepackage[table,xcdraw]{xcolor}

\theoremstyle{thmstyleone}%
\theoremstyle{thmstyletwo}%

\theoremstyle{thmstylethree}%

\renewcommand\appendixname{Supplementary Information}

\begin{document}

\title[]{Augmented CARDS: A machine learning approach to identifying triggers of climate change misinformation on Twitter}


\author[1]{\fnm{Cristian} \sur{Rojas}}\email{Cristian.RojasCardenas@monash.edu}

\author[2]{\fnm{Frank} \sur{Algra-Maschio}}\email{Frank.Algra-Maschio@monash.edu}

\author[3,6]{\fnm{Mark} \sur{Andrejevic}}\email{Mark.Andrejevic@monash.edu}

\author[4]{\fnm{Travis} \sur{Coan}}\email{T.Coan@exeter.ac.uk}

\author*[5]{\fnm{John} \sur{Cook}}\email{jocook@unimelb.edu.au}

\author*[1,6]{\fnm{Yuan-Fang} \sur{Li}}\email{Yuanfang.Li@monash.edu}

\affil[1]{\orgdiv{Department of Data Science \& AI}, \orgname{Monash University}, \orgaddress{\city{Clayton}, \postcode{3800}, \state{Victoria}, \country{Australia}}}

\affil[2]{\orgdiv{School of Social and Political Sciences}, \orgname{Monash University}, \orgaddress{\city{Clayton}, \postcode{3800}, \state{Victoria}, \country{Australia}}}

\affil[3]{\orgdiv{School of Media, Film, and Journalism}, \orgname{Monash University}, \orgaddress{\city{Clayton}, \postcode{3800}, \state{Victoria}, \country{Australia}}}

\affil[4]{\orgdiv{Exeter Q-Step Centre}, \orgname{University of Exeter}, \orgaddress{\city{Exeter}, \country{UK}}}

\affil[5]{\orgdiv{Melbourne Centre for Behaviour Change}, \orgname{University of Melbourne}, \orgaddress{\city{Parkville}, \state{Victoria}, \country{Australia}}}

\affil[6]{\orgdiv{Monash Data Futures Institute},
\orgname{Monash University},
\orgaddress{\city{Clayton}, \postcode{3800}, \state{Victoria}, \country{Australia}}}


\abstract{
Misinformation about climate change poses a significant threat to societal well-being, prompting the urgent need for effective mitigation strategies. However, the rapid proliferation of online misinformation on social media platforms outpaces the ability of fact-checkers to debunk false claims. Automated detection of climate change misinformation offers a promising solution. In this study, we address this gap by developing a two-step hierarchical model—the Augmented CARDS model—specifically designed for detecting contrarian climate claims on Twitter. Furthermore, we apply the Augmented CARDS model to five million climate-themed tweets over a six-month period in 2022. We find that over half of contrarian climate claims on Twitter involve attacks on climate actors or conspiracy theories. Spikes in climate contrarianism coincide with one of four stimuli: political events, natural events, contrarian influencers, or convinced influencers. Implications for automated responses to climate misinformation are discussed.
}




%


\maketitle

\section{Introduction}\label{sec1}
Misinformation about climate change causes a number of negative impacts. It reduces public support for mitigation policies \cite{ranney2016climate} and thwarts efforts to communicate accurate information \cite{van2017inoculating}. Misconceptions about the prevalence of contrarian views have a self-silencing effect \cite{geiger2016climate}. While misinformation has an overall impact of reducing climate literacy \cite{ranney2016climate}, this effect varies across the political spectrum, resulting in exacerbated polarisation \cite{cook2017neutralizing}.

Social media platforms have become an active site for the spread of misinformation on a wide range of topics and have received increased scrutiny for their role in undermining trust in scientific and journalistic expertise \cite{ross2022sa}. At the same time, these platforms are becoming an increasingly significant source of news and information that have an important role in shaping public awareness and discussion of issues of social importance \cite{allcott2017social}. The decentralized and networked character of the internet lowers the barriers to posting and sharing misinformation, which ends up being further amplified by engagement-maximizing commercial algorithms \cite{martens2018digital}. Regulatory regimes that protect social media platforms from editorial responsibility contribute to the “wild west” information environment in which contrarian claims circulate alongside and often more widely than traditional forms of journalistic and scientific consensus \cite{allcott2017social}. Social media also serves as a conduit for mainstreaming contrarian claims when their prevalence online results in their being taken up by news outlets and political actors \cite{tsfati2020causes}. The problems caused by the spread of misinformation online are likely to be exacerbated by recent advances in generative artificial intelligence. As an executive of a company that tracks misinformation online put it, “Crafting a new false narrative can now be done at dramatic scale, and much more frequently — it’s like having A.I. agents contributing to disinformation” \cite{Hsu2023}. 

Climate change has long been a key target of misinformation on social media platforms. Analysis of tweets around COP climate summits found that contrarian tweets and polarization have significantly grown since 2009 \citep{falkenberg2022growing}. Geographically, hoax-themed tweets that question the reality of climate change are more prominent in conservative U.S. states, relative to liberal states or tweets from the UK, Canada, or Australia \citep{jang2015polarized}. An analysis of tweets about a 2013 Intergovernmental Panel on Climate Change (IPCC) report found that Twitter users unsupportive of climate science or policies were most active in sending tweets about the IPCC, with uncivil tweets being the most viral contrarian tweets \citep{pearce2014climate}. Similarly, tweets that are skeptical about climate change have been found to show tones of incivility \citep{anderson2017social}, and climate change deniers on Twitter use aggressive language and negative sentiment \citep{effrosynidis2022exploring}.

As automated systems contribute to the generation and circulation of contrarian claims, there will be an increased need for automated detection, tracking, and response. One result will be increased pressure on journalists, platforms, watchdogs, and regulators to find ways of keeping pace with the spread of such claims. For the purposes of addressing the challenges posed by contrarian information, it is useful to be able to determine the nature of false claims. Doing so makes it possible to provide a response that addresses the substance of the claim. The ability to identify and categorize claims also makes it possible to determine the prevalence of different types of misinformation in order to shape "pre-bunking" strategies for inoculating the public against particular categories of false claims \cite{van2017inoculating}. 

It is imperative that interventions are developed and deployed to counter these negative impacts. However, this is made challenging by the fact that misinformation spreads through social media faster than factual information \cite{vosoughi2018spread}. Further, once misinformation has taken hold, it is difficult to dislodge—a phenomenon known as the continued influence effect \cite{ecker2010explicit}. Consequently, solutions that can detect and respond to misinformation in a rapid fashion are required.

However, automatic detection and correction of misinformation are technically challenging, earning the label "the holy grail of fact-checking" \cite{hassan2015quest}. There have been efforts to automatically detect and fact-check misinformation across various domains \citep{andersen2020communicative,guo2022survey}. On climate misinformation, there have been few efforts to detect misinformation. Unsupervised topic analysis has been employed to identify the major themes in conservative think-tank (CTT) texts \cite{boussalis2016text}, link corporate funding to polarizing climate text \citep{farrell2016corporate}, and identify climate framings in newspaper articles \citep{stecula2019framing}. There have also been efforts to detect logical fallacies in climate misinformation as well as across general topics \cite{alhindi2023multitask, jin2022logical,zanartu2023fallacy}.

The CARDS (Computer Assisted Recognition of Denial \& Skepticism) model used supervised machine learning to detect and categorize contrarian claims about climate change \cite{coan2021computer}. The model has been shown to be effective in categorising a wide range of contrarian claims about climate change. The model was based on a taxonomy of contrarian claims consisting of five main categories: 1) global warming isn't happening, 2) humans aren't causing global warming, 3) climate impacts aren't bad, 4) climate solutions won't work, and 5) climate movement/science are unreliable). At the second level of this taxonomy are sub-categories of contrarian claims such as 5.2 (climate actors are unreliable) and 5.3 (conspiracy theories).

\begin{figure}[h!]
    \centering
    \includegraphics[width=0.9\textwidth]{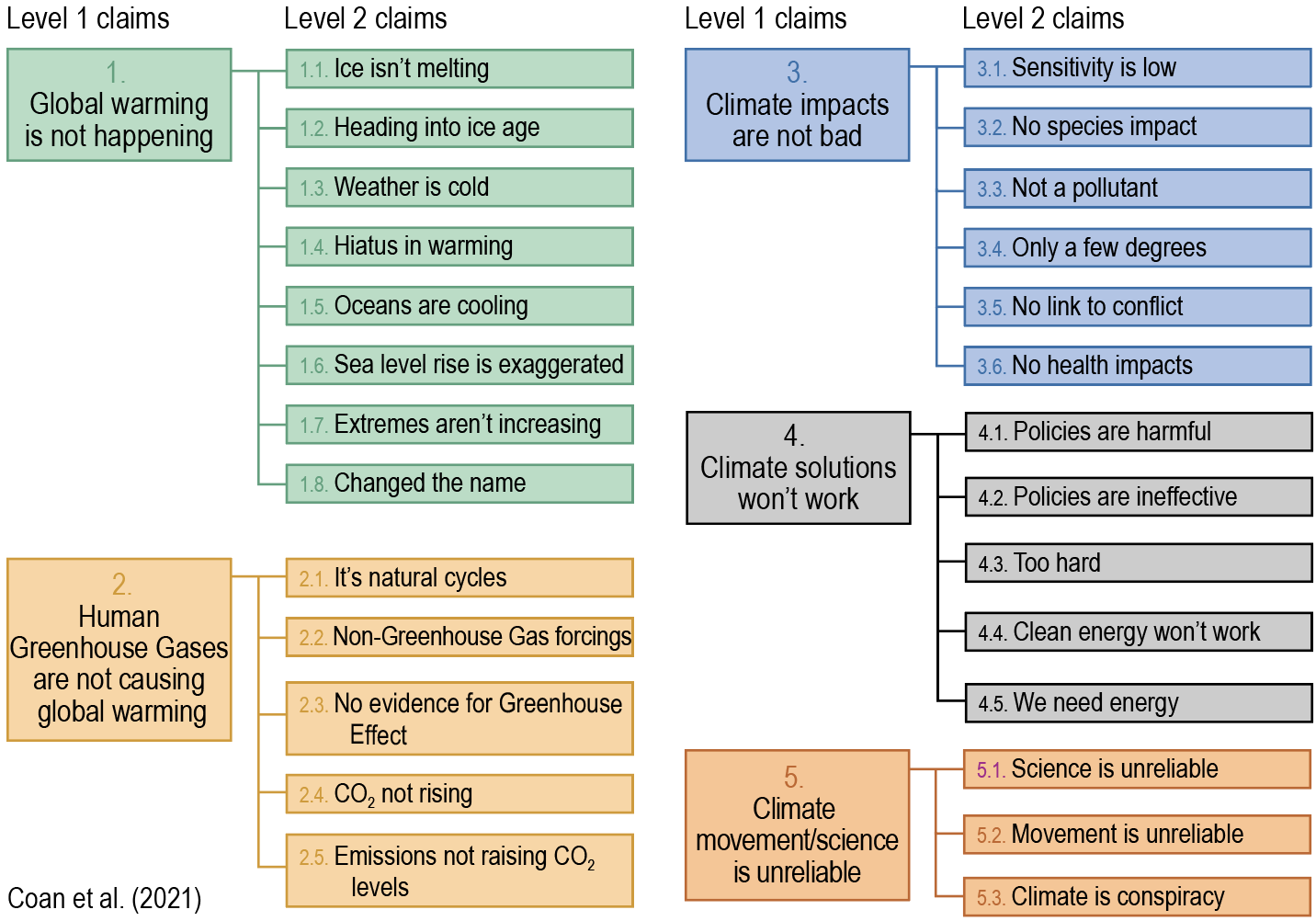}
    \caption{\textbf{CARDS Taxonomy of Contrarian Climate Claims} This taxonomy provides a comprehensive overview of the frequently employed main claim and its corresponding subarguments utilized to bolster contrarian perspectives on climate change.\citep{coan2021computer}.}
    \label{fig:timelines}
\end{figure}

However, the CARDS model was only trained using text from contrarian blogs and conservative think-tank websites—prolific sources of climate misinformation—and its performance in classifying content from other datasets (e.g., from social media platforms) has yet to be assessed. This study assesses and augments the CARDS model's performance in classifying contrarian climate claims in Twitter data. We apply the Augmented CARDS model to a dataset of climate tweets, in order to examine the various arguments that are characteristic of different types of contrarian peaks.

\section{Methods}\label{sec11}

The original CARDS model was trained using a dataset comprising paragraphs extracted from sources known for their wealth of climate contrarian content, such as conservative think-tank articles and contrarian blog posts. This training approach showed strong performance when tested on similar content sources. Nevertheless, the model's ability to effectively differentiate between contrarian and convinced text (reflecting the scientific consensus on climate change) within the context of Twitter remained uncertain. To mitigate this uncertainty, we present an enhanced CARDS model introducing an initial binary classifier. This classifier's primary function is to distinguish between convinced and contrarian claims, aided by the inclusion of supplementary Twitter data. Subsequently, we include an additional layer responsible for classifying contrarian claims into their respective typology.

\subsection{Model Architecture}

Augmented CARDS enhances the performance of the original CARDS model on Twitter by utilizing additional data from the platform and rectifying category imbalances through a two-stage hierarchical architecture. The general model architecture, consist of an initial layer trained in a binary detection task to differentiate between convinced and contrarian tweets, coupled with an additional layer trained in a multilabel task to classify the taxonomy.

\begin{figure}[h!]
    \centering
    \includegraphics[width=0.9\textwidth]{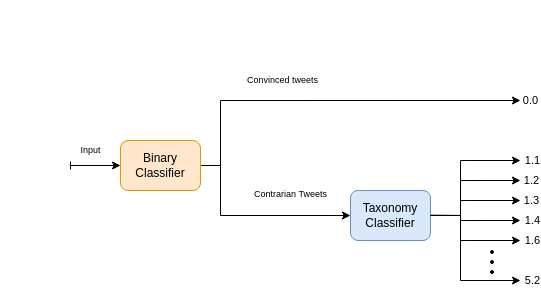}
    \caption{\textbf{Model Architecture}}
    \label{fig:architecture}
\end{figure}

Both classifiers incorporate the DeBERTa language model, structured based on the auto-encoding transformer architecture introduced in BERT \cite{devlin-etal-2019-bert}. The innovation includes disentangled attention and a more extensive pretraining process \cite{liu2019roberta,He2020DeBERTaDB}. Specifically, we utilized the large version of DeBERTa, consisting of 24 transformer blocks with a hidden size of 1024 and 16 attention heads. Additionally, an extra dense layer was employed for the classification task, bringing the total number of parameters to approximately 355 million.

We aim to specialize the classifiers in their respective tasks, undergoing training tailored to their specific contexts. The implementation of a hierarchical architecture responds to the necessity of modularizing both tasks to effectively handle the fine-tuning process of the pipeline and improve its performance. Additionally, it mitigates the issue of unbalanced data distribution. Given that the datasets are overly dominated by convinced claims, the challenge lies in effectively detecting the remaining 17 classes of the taxonomy.

Moreover, DeBERTa's transfer learning capabilities are mainly attributed to its pretraining on web-sourced texts. Nevertheless, since Twitter was not incorporated into the pretraining procedure \cite{He2020DeBERTaDB}, fine-tuning is necessary to capture the linguistic features specific to the platform.

\subsection{Training Details}

To enhance the model's performance, we incorporated the Climate Change Twitter Dataset labeled by the University of Waterloo, featuring a 90/10 ratio of verified and misleading tweets, \cite{waterloodataset} to the binary classifier training set. Furthermore, the taxonomy classifier underwent training using the CARDS dataset, incorporating the 5.3 category ("climate change is a conspiracy theory"), which differed from original CARDS which merged category 5.3 with category 5.2. Separating these two categories was deemed appropriate due to the substantial prevalence of conspiracy theories in climate change tweets.

The models were fine-tuned over 3 epochs with a learning rate of 1e-5 in a v100 GPU with a batch size of 6. The input was constrained to sequences of 256 tokens with a padding method. These parameters, along with the seed were kept constant for comparison with the original CARDS method.

To assess the model's capabilities, climate change experts labelled a testing set of tweets following the \cite{coan2021computer} taxonomy. This dataset, denoted as \textit{``Expert Annotated Climate Tweets"} in Table \ref{table:hierarchicalresults}, was composed of 2607 tweets related to climate change, sampled from the platform in the second half of 2022.

\subsection{Data Analysis}

The analysis was carried out on a large dataset of climate change-related tweets. This dataset was compiled between July and December of 2022 by the Online Media Monitor (OMM) at the University of Hamburg \cite{omm2023}. We examined the temporal frequency of the data and identified intervals of interest based on distinct patterns discerned by the model. Within these intervals, a word frequency analysis was conducted and compared against the overall word frequency of the entire dataset. This comparison enabled us to highlight specific shifts in word usage during those periods and establish a connection between this information and relevant events that took place.

The word frequency was calculated by comparing the log-fold change and the p-value derived from the distribution differences for various words. Subsequently, a filter was applied to keep only those words with a significance level greater than 0.05, and they were ranked based on their log-fold change in descending order. Finally, the top 10 most relevant words were used to characterize the event (see Supplementary Table \ref{appendice:words}).

\subsection*{Data availability}

The experimental data used to train the models and perform the analysis in this study are available on Figshare, accessible through the following identifier: 
\href{https://doi.org/10.6084/m9.figshare.25465036}{https://doi.org/10.6084/m9.figshare.25465036}.

\subsection*{Code availability}

The machine learning models used for the analysis conducted in this study can be found in the following repositories for reproducibility of our results:
\begin{enumerate}
    \item \href{https://huggingface.co/crarojasca/BinaryAugmentedCARDS}{https://huggingface.co/crarojasca/BinaryAugmentedCARDS}.

    \item \href{https://huggingface.co/crarojasca/TaxonomyAugmentedCARDS}{https://huggingface.co/crarojasca/TaxonomyAugmentedCARDS}.
\end{enumerate}

\section{Results}\label{sec2}

\subsection{Assessing the Augmented CARDS model}

Table \ref{table:hierarchicalresults} compares the performance of the original CARDS and  Augmented CARDS models in identifying contrarian claims in the original CARDS testing set (comprised of contrarian blogs and CTTs) and in our new Twitter dataset. We subdivide this task into two stages: binary detection (distinguishing between contrarian and convinced claims) and taxonomy detection (identifying claims from the CARDS taxonomy).

The original CARDS model performed exceptionally well in datasets sharing linguistic features with its original training data, including CTT articles and contrarian blog posts. This is demonstrated by the F1-score achieved in CARDS for binary detection (89.9), slightly outperforming Augmented CARDS.

\begin{table}[]
    \begin{subtable}{.55\linewidth}
    \begin{tabular}{@{}
    >{\columncolor[HTML]{FFFFFF}}r 
    >{\columncolor[HTML]{FFFFFF}}r 
    >{\columncolor[HTML]{FFFFFF}}r 
    >{\columncolor[HTML]{FFFFFF}}r 
    >{\columncolor[HTML]{FFFFFF}}r @{}}
    \toprule
    \multicolumn{1}{c}{\cellcolor[HTML]{FFFFFF}} &
      \multicolumn{1}{c}{\cellcolor[HTML]{FFFFFF}} &
      \multicolumn{2}{c}{\cellcolor[HTML]{FFFFFF}\textbf{Models}} &
      \multicolumn{1}{c}{\cellcolor[HTML]{FFFFFF}} \\ \cmidrule(lr){3-4}
    \multicolumn{1}{c}{\multirow{-2}{*}{\cellcolor[HTML]{FFFFFF}\textbf{Task}}} &
      \multicolumn{1}{c}{\multirow{-2}{*}{\cellcolor[HTML]{FFFFFF}\textbf{Datasets}}} &
      \textbf{CARDS} &
      \textbf{Augmented CARDS} &
      \multicolumn{1}{c}{\multirow{-2}{*}{\cellcolor[HTML]{FFFFFF}\textbf{Support}}} \\ \midrule
    \cellcolor[HTML]{FFFFFF}                                              & \textit{CARDS}                           & \textbf{90.3} & 89.1          & 2931 \\
    \cellcolor[HTML]{FFFFFF}                                              & \textit{Twitter Climate Change}          & 68.1          & \textbf{88.2} & 4395 \\
    \multirow{-3}{*}{\cellcolor[HTML]{FFFFFF}\textit{Binary Detection}}   & \textit{Expert Annotated Climate Tweets} & 67.8          & \textbf{81.1} & 2711 \\ \midrule
    \cellcolor[HTML]{FFFFFF}                                              & \textit{CARDS}                           & \textbf{72.7} & 73.100        & 2904 \\
    \multirow{-2}{*}{\cellcolor[HTML]{FFFFFF}\textit{Taxonomy classification}} & \textit{Expert Annotated Climate Tweets} & 43.7          & \textbf{53.6} & 2607 \\ \bottomrule
    \end{tabular}
    \label{table:hierarchicalresults}
    \caption{\label{table:hierarchicalresults}}
    \end{subtable}
    
    \begin{subtable}{.55\linewidth}
        \begin{tabular}{@{}rrrr@{}}
        \toprule
        \rowcolor[HTML]{FFFFFF} 
        \textbf{Category}      & \textbf{CARDS}       & \textbf{Augmented CARDS} & \textbf{Support}     \\ \midrule
        \rowcolor[HTML]{FFFFFF} 
        0.0 & 70.9          & \textbf{81.5} & 1049 \\
        \rowcolor[HTML]{FFFFFF} 
        1.1 & 60.5          & \textbf{70.4} & 28   \\
        \rowcolor[HTML]{FFFFFF} 
        1.2 & 40            & \textbf{44.4} & 20   \\
        \rowcolor[HTML]{FFFFFF} 
        1.3 & 37            & \textbf{48.6} & 61   \\
        \rowcolor[HTML]{FFFFFF} 
        1.4 & 62.1          & \textbf{65.6} & 27   \\
        \rowcolor[HTML]{FFFFFF} 
        1.6 & 56.7          & \textbf{59.7} & 41   \\
        \rowcolor[HTML]{FFFFFF} 
        1.7 & 46.4          & \textbf{52}   & 89   \\
        \rowcolor[HTML]{FFFFFF} 
        2.1 & 68.1          & \textbf{69.4} & 154  \\
        \rowcolor[HTML]{FFFFFF} 
        2.3 & \textbf{36.7} & 25            & 22   \\
        \rowcolor[HTML]{FFFFFF} 
        3.1 & \textbf{38.5} & 34.8          & 8    \\
        \rowcolor[HTML]{FFFFFF} 
        3.2 & 61            & \textbf{74.6} & 31   \\
        \rowcolor[HTML]{FFFFFF} 
        3.3 & 54.2          & \textbf{65.4} & 23   \\
        \rowcolor[HTML]{FFFFFF} 
        4.1 & 38.5          & \textbf{49.4} & 103  \\
        \rowcolor[HTML]{FFFFFF} 
        4.2 & \textbf{37.6} & 28.6          & 61   \\
        \rowcolor[HTML]{FFFFFF} 
        4.4 & 30.8          & \textbf{54.5} & 46   \\
        \rowcolor[HTML]{FFFFFF} 
        4.5 & 19.7          & \textbf{39.4} & 50   \\
        \rowcolor[HTML]{FFFFFF} 
        5.1 & 32.8          & \textbf{38.2} & 96   \\
        \rowcolor[HTML]{FFFFFF} 
        5.2 & 38.6          & \textbf{53.5} & 498  \\
        \rowcolor[HTML]{FFFFFF} 
        5.3 & -             & \textbf{62.9} & 200  \\
        \multicolumn{1}{l}{}   & \multicolumn{1}{l}{} & \multicolumn{1}{l}{}     & \multicolumn{1}{l}{} \\
        \rowcolor[HTML]{FFFFFF} 
        \textbf{Macro Average} & 43.69                & \textbf{53.57}           & 2407                 \\ \bottomrule
        \end{tabular}
        \label{table:taxonomyscores}
        \caption{\label{table:taxonomyscores} }
    \end{subtable}
    \caption{ \textbf{Machine Learning Model Metrics.}  (\subref{table:hierarchicalresults}). Assessment of F1-scores achieved, comparing the original CARDS model with the Augmented CARDS Model. (\subref{table:taxonomyscores}) F1-scores per category obtained from the Augmented CARDS model on the \textit{``Expert Annotated Climate Tweets"} dataset. Bold values indicate which model showed higher performance.}
    \label{table:modelresults}
\end{table}

However, in the taxonomy detection task, the original model showed a 5\% performance decrease relative to the CARDS metrics \cite{coan2021computer}. This decline is attributed to the inclusion of the 5.3 category (contrarian claims involving conspiracy theories) in our analysis. This category is highly relevant in the Twitter context but was excluded from the original model in \cite{coan2021computer}. In this scenario, the Augmented CARDS architecture demonstrated better adaptability, achieving a 76.6 F1-score with additional data from Twitter, where climate change conspiracy arguments hold more significance among contrarians.

Nonetheless, our results indicate that Twitter is a challenging task due to the significant disparities in language and writing style observed between the original sources and the platform. On the other hand, the Augmented CARDS model achieves a significant improvement in the F1-Score for both the \textit{``Twitter Climate Change"} and the \textit{``Expert Annotated Climate Tweets"} datasets for both tasks.

The technical advantages of Augmented CARDS included leveraging additional data from the Twitter context and addressing category imbalances through a hierarchical architecture. Based on these two factors, as shown in Table \ref{table:hierarchicalresults}, the Augmented CARDS model demonstrated a relative 16\% performance improvement for binary detection and 14.3\% for taxonomy detection on our \textit{``Expert Annotated Climate Tweets"} dataset. This translates to an F1-score of 81.6 for binary detection and 53.4 for taxonomy detection, while maintaining a similar level of performance in the original domain. Although there is still room for improvement, especially in taxonomy detection, it would require collecting a larger Twitter-based dataset for the less common categories in this context. Most of the categories with low F1-scores are infrequent on Twitter as illustrated in Table \ref{table:taxonomyscores}.

In contrast, on Twitter, the most prominent contrarian categories were 5.2 (climate actors are unreliable), 5.3 (conspiracy theories), 4.1 (policies are harmful), 2.1 (global warming is naturally caused), and 1.7 (extreme weather isn't linked to climate change). Table \ref{table:taxonomyscores} shows that our model exhibits the most substantial improvements with these categories. The F1-scores achieved by Augmented CARDS demonstrate an overall enhancement across most categories, with major improvement in the more relevant ones. Compared to blogs and CTT articles, the distribution of contrarian arguments on Twitter shows a different emphasis, with ad hominem fallacies (category 5.2) directed at climate actors being the most common type of argument. The second most common type of contrarian argument is conspiracy theories about climate change (category 5.3).

\subsection{Applying Augmented CARDS to 2022 climate tweets}

We applied the Augmented Cards model to over 5 million climate-related tweets in a six-month period in 2022, providing insight into the proliferation of climate-contrarian claims on Twitter. This novel investigation enabled an analysis of the triggers that caused an upsurge in contrarian claims on the platform and the most common types of contrarian claims.

The tweets used in our analyses were collected by the University of Hamburg by filtering for terms similar to "climate change" \cite{omm2023} (see Supplementary Data \ref{appendice:data}). Figure \ref{fig:timeline} illustrates the daily frequency of tweets related to climate change, showing notable fluctuations in the frequency of climate tweets, such as the significant peak in late July. On average, 27,464 tweets per day are related to climate change in this data set, with the significant peaks in late July and mid-November resulting in 65,196 and 43,647 of tweets respectively. 

\begin{figure}[t]
    \centering
    \begin{subfigure}{\textwidth}
    \centering
        \includegraphics[width=0.9\textwidth]{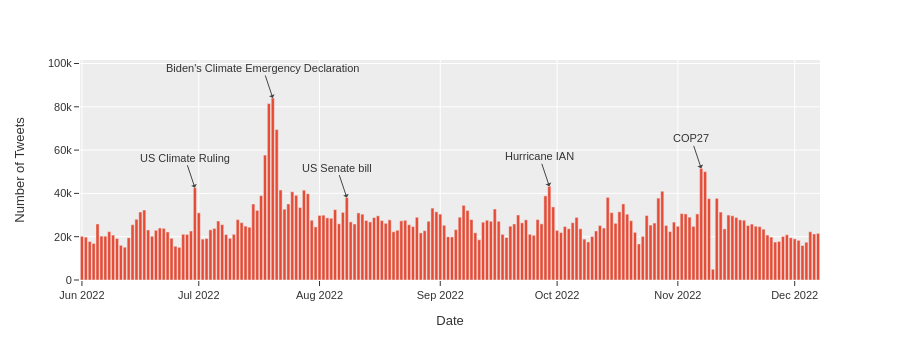}
        \label{fig:timeline}
        \caption{\label{fig:timeline} }
    \end{subfigure}

    \begin{subfigure}{\textwidth}
    \centering
        \includegraphics[width=0.9\textwidth]{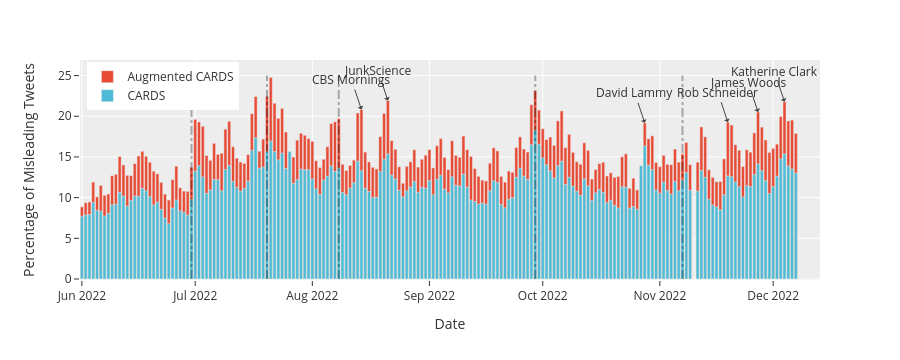}
        \label{fig:misinformation}
        \caption{\label{fig:misinformation}}
    \end{subfigure}
    \caption{\textbf{Climate change Tweeter trends in 2022} (\subref{fig:timeline})  Number of tweets related to climate change topics by date in the 2022 period. (\subref{fig:misinformation}) Percentage of misinformative tweets detected by the CARDS and Augmented CARDS models.}
    \label{fig:timelines}
\end{figure}

To investigate attributes of these peaks, we performed statistical analyses to identify words with major variations and establish correlations between these shifts and significant events that occurred during the corresponding periods. The word frequency analysis involved comparing changes in word distributions during specific periods in relation to the entire dataset. We computed the log fold change and p-value to assess differences in these distributions (see Supplementary Table \ref{appendice:words} for more details).

Between July 19 and 21, marking the period with the largest peak in climate tweets, the terms  "climate emergency" and "Biden" showed the greatest shifts. Based on news reports from that time, these discussions occurred when it became apparent that President Joe Biden was considering declaring a climate emergency in response to the heatwave affecting both the United States and Europe \cite{Smith.2022}. 

The second-largest peak of overall climate tweets was associated with  COP27, as indicated by the changes in word distribution illustrated in Table \ref{table:dwordanalysis}. This event was associated with a doubling of the number of tweets between September 7th and 9th, 2022. The third highest peak of overall climate tweets in our dataset coincided with Hurricane Ian \cite{RichardLuscombe}. Tweets relating to the Hurricane became the major topic of discussion related to climate change between September 28 and October 1, although they generated only half the number of tweets compared to Biden's declaration event.

Turning now to the analysis of contrarian tweets, Figure \ref{fig:misinformation} displays the percentage per day of contrarian tweets detected by the Augmented CARDS model through a binary inference process. This analysis indicates that the average proportion of contrarian tweets per day is 15.5\%, yet there is clearly a number of peaks of contrarian tweets throughout the six-month period. 

Overall, we identified four distinct categories of events that coincided with an upsurge in the publication of contrarian tweets, as outlined in Table \ref{table:event_types}. The triggering events can be broadly classified into three primary groups: Natural Events, Political Events, and Influencer Posts.

\begin{table}[h]
\begin{tabular}{@{}cl@{}}
\toprule
\rowcolor[HTML]{FFFFFF} 
\textbf{Nature of trigger} & \multicolumn{1}{c}{\cellcolor[HTML]{FFFFFF}\textbf{Events}} \\ \midrule
\rowcolor[HTML]{FFFFFF} 
\textit{Natural Event}     & - Hurricane IAN                                             \\
\hline
\rowcolor[HTML]{FFFFFF} 
\textit{Political Event}       & \begin{tabular}[c]{@{}l@{}}- US Climate Ruling\\ - Biden's Climate Emergency Declaration\\ - US Senate bill\end{tabular} \\
\hline
\rowcolor[HTML]{FFFFFF} 
\textit{Contrarian Influencer} & \begin{tabular}[c]{@{}l@{}}- Steve Milloy\\ - Rob Schneider\\ - James Woods\end{tabular}                                 \\
\hline
\rowcolor[HTML]{FFFFFF} 
\textit{Convinced Influener}   & \begin{tabular}[c]{@{}l@{}}- Dan Rather\\ - CBS Mornings\\ - David Lammy\\ - Katherine Clark\end{tabular}                \\ \bottomrule
\end{tabular}
\caption{ List of occurrences that induced spikes in climate contrarianism on Twitter.}
\label{table:event_types}
\end{table}

Natural Events, such as Hurricane IAN, and Political Events like COP27, were external occurrences originating outside the platform \cite{Luscombe_2022,Zee_Horton_2022}. They coincided with a general increase in public discourse surrounding the climate change topic and occasionally prompted shifts in contrarian positions.

\begin{figure*}[]
    \centering
    \begin{subfigure}[t]{0.5\textwidth}
        \includegraphics[height=1.6in]{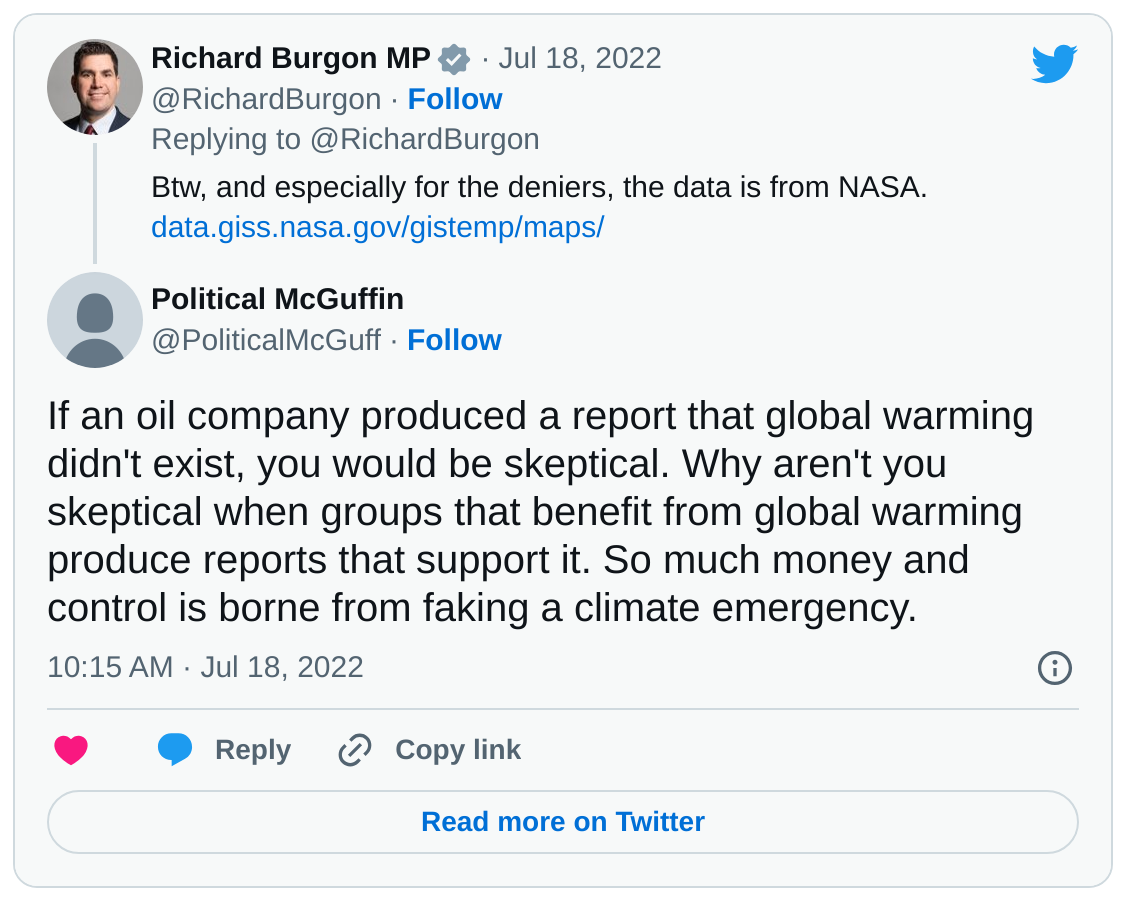}
    \end{subfigure}%
    ~ 
    \begin{subfigure}[t]{0.5\textwidth}
        \includegraphics[height=1.4in]{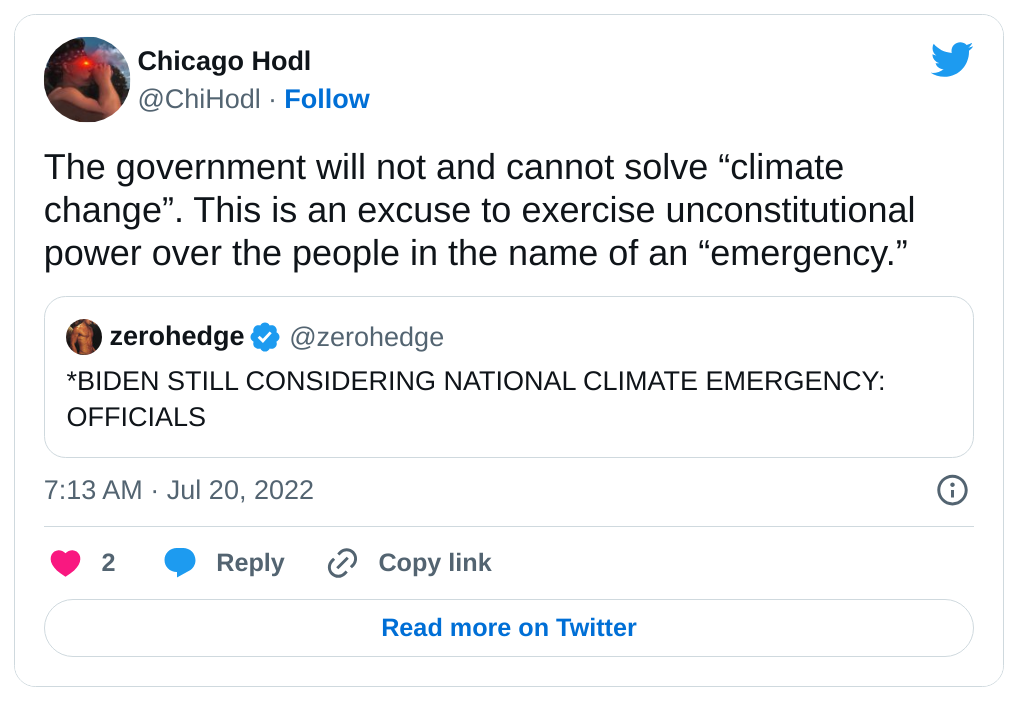}
    \end{subfigure}
    \caption{Tweets sampled from the trend peak related to climate change observed between July 18 and July 21, 2022.}
    \label{fig:tweet_5.2}
\end{figure*}

For example, the Biden declaration was seized upon by climate change contrarians, triggering significant peaks in the percentage of contrarian tweets. In Figure \ref{fig:tweet_5.2}, we present several examples illustrating some of the contrarian opinions. The primary concern revolved around the possibility that climate warming might be used as a political pretext to declare an emergency, potentially granting expanded powers to President Biden, which could disrupt the existing state equilibrium. This event caused the percentage of contrarian claims to surpass 20\%, reaching a peak of 24.7\%. 

\begin{figure}[h!]
    \centering
    \begin{subfigure}[t]{0.5\textwidth}
        \includegraphics[height=1.2in]{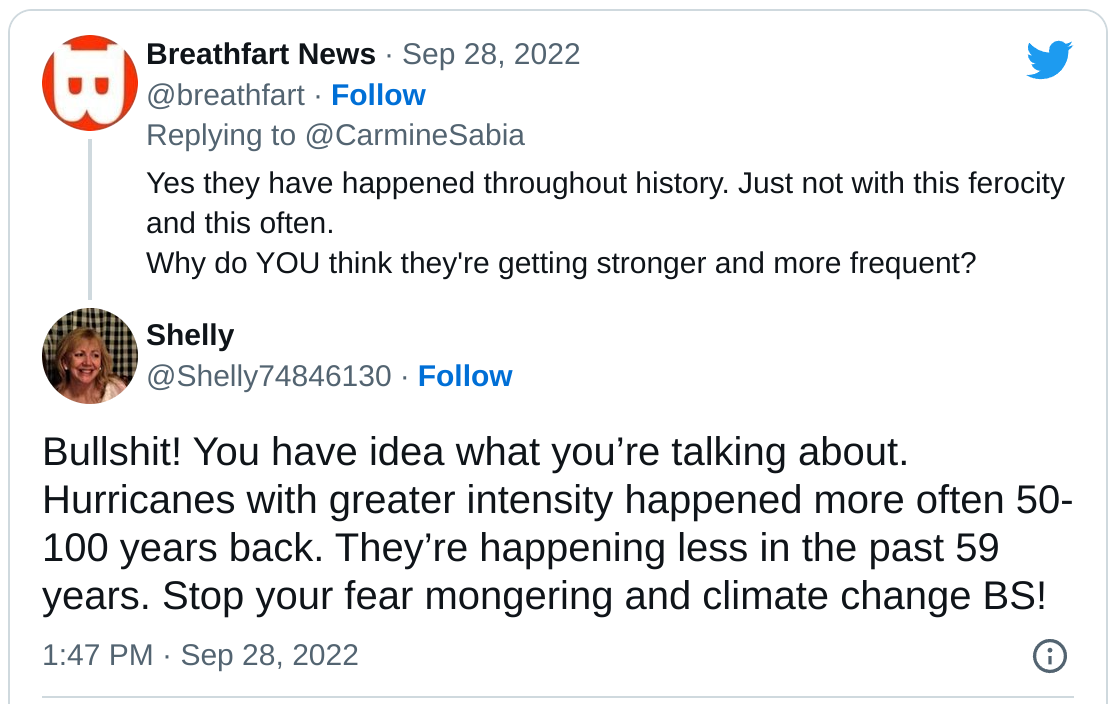}
    \end{subfigure}%
    ~ 
    \begin{subfigure}[t]{0.5\textwidth}
        \includegraphics[height=1.6in]{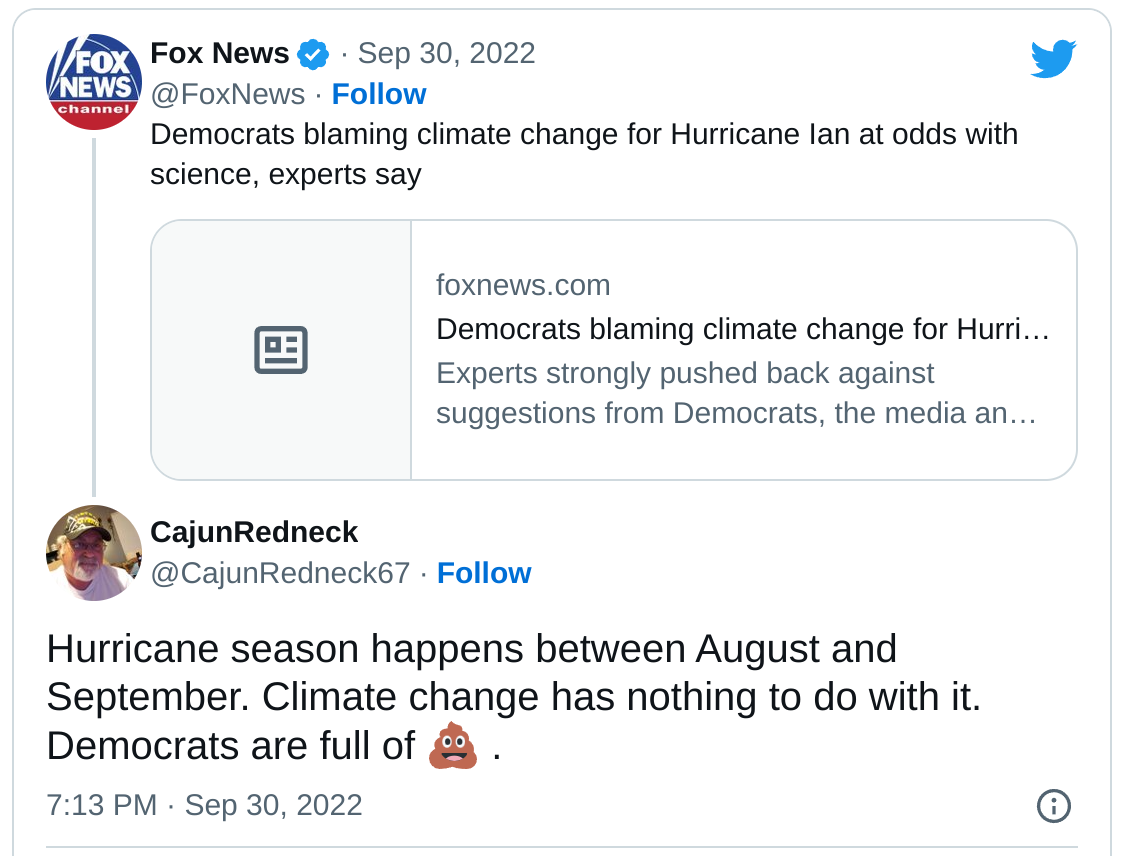}
    \end{subfigure}
    \caption{Tweets that deny the link between extreme weather and climate change.}
    \label{fig:tweet_1.7}
\end{figure}

Similarly, the Natural Event of Hurricane Ian triggered an increase in all tweets related to climate change and the percentage of contrarian claims. Despite generating only half the number of tweets compared to Biden's declaration, the proportion of contrarian tweets related to Hurricane Ian reached similarly high levels. Discussions were centred around the impact of climate change on extreme weather conditions. Some examples of tweets are illustrated in Figure \ref{fig:tweet_1.7}. For instance, FoxNews tweeted one of their articles, titled "Democrats blaming climate change for Hurricane Ian at odds with science, experts say". 

The COP27 event highlights a contrast between the Biden emergency declaration and Hurricane Ian. While both of these latter events led to a rise in the volume of tweets and contrarian claims, COP27 triggered only a 1 per cent increase in the proportion of contrarian claims among climate tweets. 

While Political and Natural Events coincided with an increase in the volume of tweets related to climate change and, in some cases the percentage of contrarian claims, there was no significant increase in the overall volume of climate tweets associated with Influencer Posts. Further, contrarian claims increased in response to influencers regardless of whether they expressed a convinced or contrarian view. These influencers could be politicians, comedians, film directors, or media figures. Nonetheless, they all share the characteristic of being public figures with a substantial number of followers. It's important to note that our categorization of influencers is based on the positions adopted by their publications during 2022, not necessarily their current personal viewpoints.

\begin{figure}[t]
    \centering
    \includegraphics[height=2.5in]{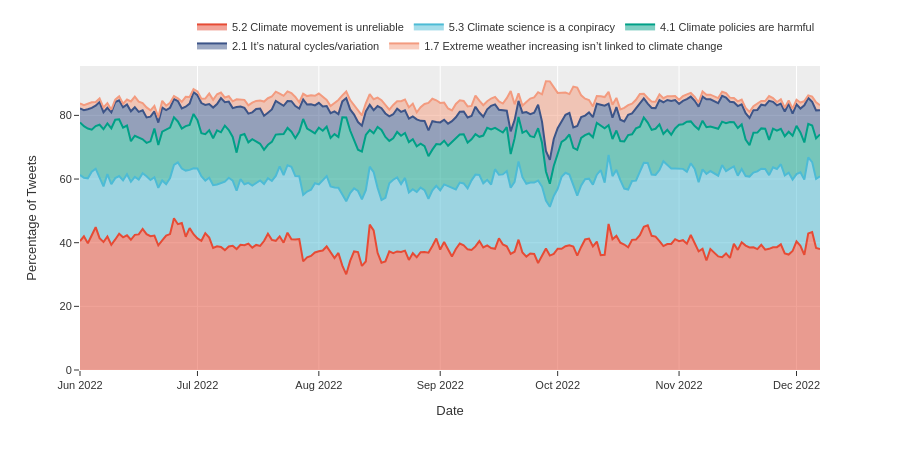}
    \caption{\textbf{Distribution of the Top 5 Contrarian Arguments} Breakdown of the five most relevant categories detected by the Augmented CARDS model on Tweeter in 2022.}
    \label{fig:miinfopattern}
\end{figure}

Our final analysis is the categorisation of contrarian tweets by the typology of \cite{coan2021computer} as inferred by the Augmented CARDS model. Figure \ref{fig:miinfopattern} represents the distribution of the tweets by the most common categories identified in the climate-related tweets. The distribution of contrarian categories remains relatively stable even on dates with significant deviations. The most common form of climate contrarianism involves criticisms of climate actors such as climate scientists and environmentalists (category 5.2), comprising 40\% of the total number of misleading tweets. This is followed by category 5.3, which includes tweets categorizing climate change as a conspiracy, making up approximately 20\% of the segment. Categories 4.1 (climate policies are harmful) and 2.1 (natural cycles are causing global warming, not humans) make up the next two most relevant categories. The fifth most common category, 1.7 (extreme events are not increasing) receives a significant share of the distribution during the Hurricane Ian period. 

\begin{table}[t]
\begin{tabular}{@{}crrrrrr@{}}
\toprule
Nature of trigger &
  \multicolumn{1}{c}{\textbf{5.2}} &
  \multicolumn{1}{c}{\textbf{5.3}} &
  \multicolumn{1}{c}{\textbf{4.1}} &
  \multicolumn{1}{c}{\textbf{2.1}} &
  \multicolumn{1}{c}{\textbf{1.7}} &
  \multicolumn{1}{c}{\textbf{others}} \\ \midrule
\textit{Contrarian Influencer} & -2    & \textbf{12.84} & -17.19        & 5.82           & -41.19          & 9.29   \\
\textit{Convinced Influener}   & 3.07  & 9.1            & -15.37        & \textbf{11.05} & -35.96          & -4.13  \\
\textit{Natural Event}         & -8.24 & -26.76         & -48.77        & 0.37           & \textbf{680.15} & -38.22 \\
\textit{Political Event}       & -3.24 & 4.23           & \textbf{25.2} & 2.15           & -31.55          & -15.62 \\ \bottomrule
\end{tabular}
\caption{\textbf{Preferred arguments based on the type of trigger} Percent changes in the distribution of contrarian tweets based on the nature of the trigger. Bold values indicate the most frequently used category as an argument by each trigger.}
\label{table:percertual_increase_category}
\end{table}

Generating a time evolution of contrarian claims allows us to identify which categories dominate based on the different types of triggers. Table \ref{table:percertual_increase_category} shows that Natural Events and Political Events shifted the distribution towards topics related to categories 1.7 and 4.1, respectively. This is expected given that 1.7 relates specifically to extreme events and increased during the Hurricane Ian period. Moreover, increases in category 4.1 were associated with political events, which is to be expected given that the category involves criticisms of climate policies.

The fluctuations generated by influencers lean significantly towards categories 5.3 (conspiracy theories) and 2.1 (natural cycles/variation), irrespective of whether the influencer holds a contrarian or convinced stance. Notably, when the influencer supported a contrarian viewpoint, there was a discernible increase in the prevalence of conspiracy theories. Conversely, in instances where the influencer was convinced, the distribution leaned slightly more toward posts stating that climate change as a natural cycle, accompanied by a concurrent rise in conspiracy theories.

\begin{figure*}[]
    \centering
    \begin{subfigure}[t]{0.5\textwidth}
        \includegraphics[height=1.7in]{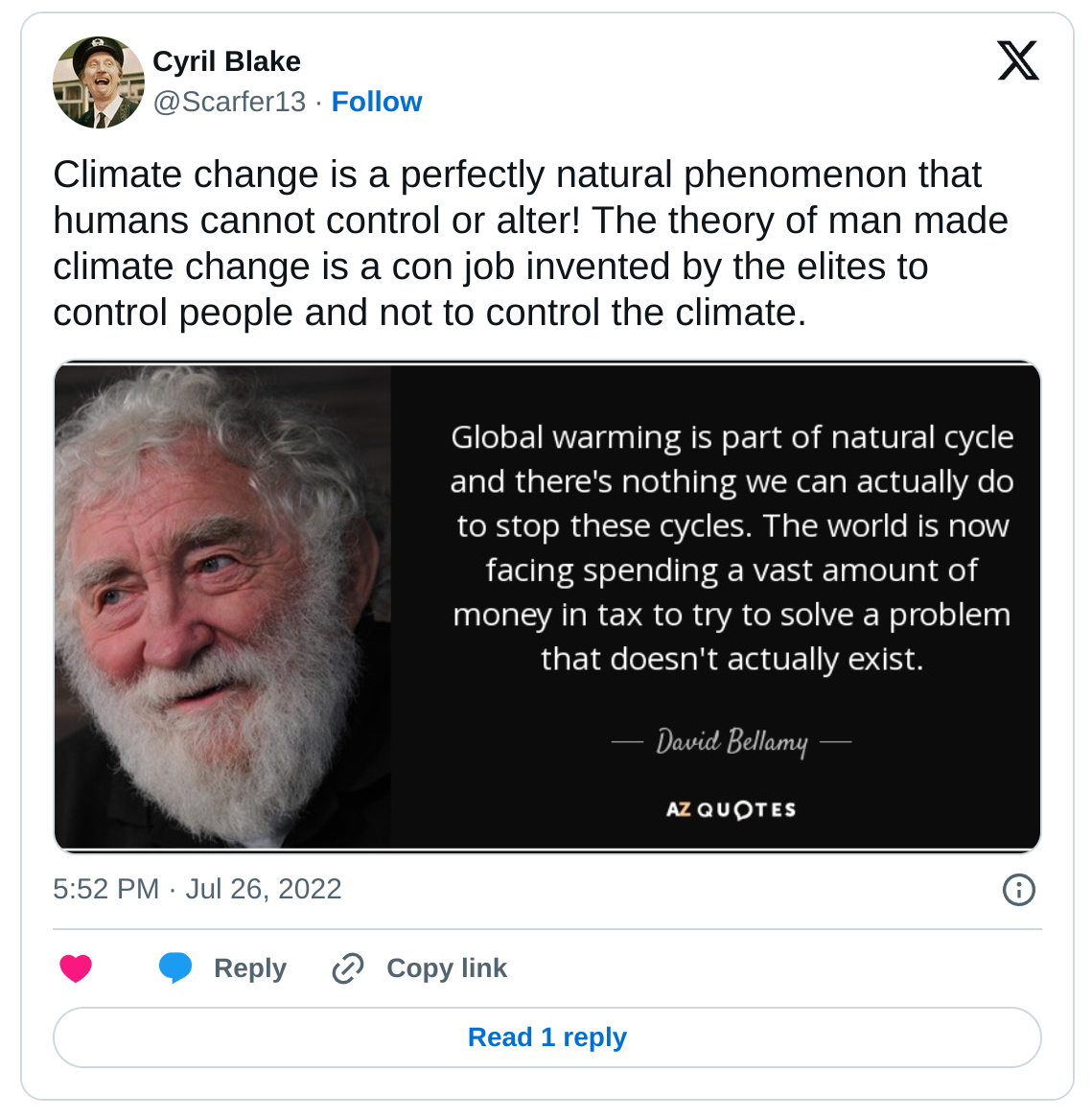}
        \label{fig:misinfoa}
        \caption{\label{fig:misinfoa} }
    \end{subfigure}%
    ~ 
    \begin{subfigure}[t]{0.5\textwidth}
        \includegraphics[height=1.3in]{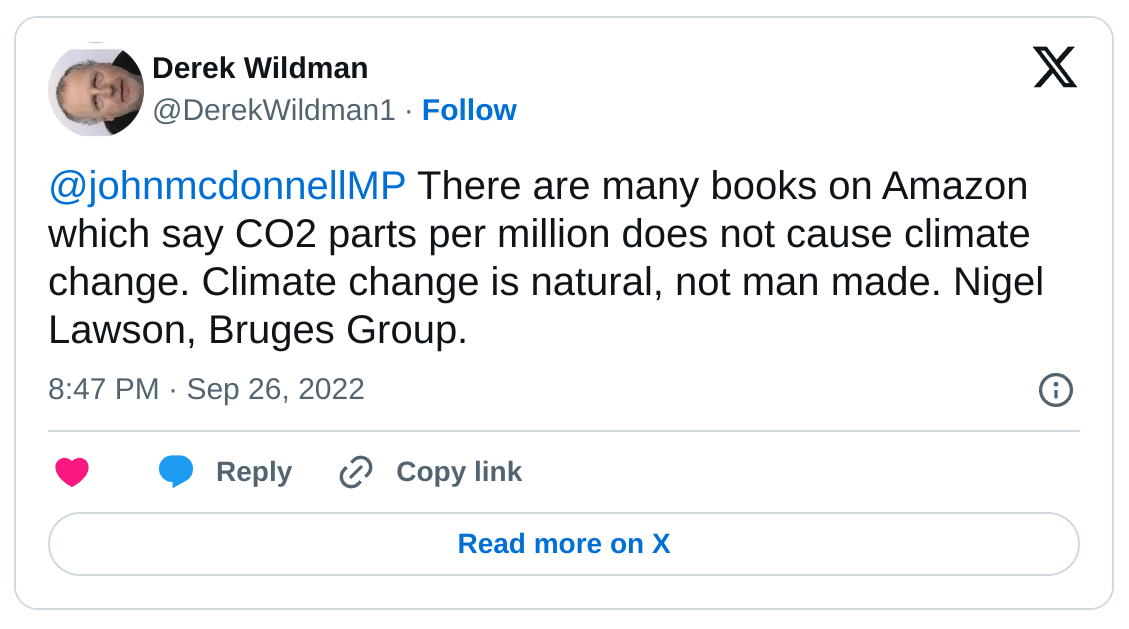}
        \label{fig:misinfob}
        \caption{\label{fig:misinfob} }
    \end{subfigure}
    \caption{Tweets sampled from outlier accounts with a high number of contrarian publications.}
    \label{fig:misinfo}
\end{figure*}

Finally, a considerable number of accounts exhibited an unusually high volume of contrarian tweets across all primary categories. On average, each user shared between 1 and 2 contrarian tweets. Notably, category 5.3 had the highest average of 1.9 tweets per user, closely followed by category 5.2 with an average of 1.8. The remaining three categories had averages of 1.57 contrarian tweets or fewer per user. This suggests that discussions revolving around conspiracy theories were generating a significant amount of content from a relatively closed pool of users. However, users with hundreds of published tweets could be found across all principal categories. For example, the user with the maximum number of posts related to conspiracies (5.3) reached 921 tweets just in this category. Similar cases could be found in other categories as well: 5.2 (881), 4.1 (627), 2.1 (424), and 1.7 (108). Some of these outlier accounts were evidently automated. For instance in Figure \ref{fig:misinfob}, one user posted 1977 tweets within our data period, with only 206 of them featuring unique textual content, while the rest repetitively rephrased the same idea with minor grammatical changes. Other accounts were evidently managed by individuals engaging in discussions and expressing opinions. Examples of such accounts can be observed in Figure \ref{fig:misinfoa}. Overall, the proportion of unique content in contrarian detections is approximately 93.3\%, with categories 5.3 and 5.2 containing around 91.3\% and 94.1\% unique content, respectively. In summary, the existence of accounts managed by automated systems and users generating a substantial volume of contrarian tweets is indisputable, with their content consistently reflecting the trends identified in our taxonomy analysis. Additionally, we observed that approximately 6\% of the analyzed tweets could be categorized as spam. However, content generated by modern AI models could potentially circumvent our current analytical approach and exacerbate this issue.

\section{Discussion}\label{sec12}

Our study showed that a classifier trained only on climate contrarian text (e.g., the original CARDS model) struggles at the binary classification task of distinguishing between convinced and contrarian text. We found that adding training data that includes annotations of both convinced and contrarian examples from improved performance in binary classification. This addressed a limitation of the original CARDS model, which performed well with known misinformation sources but struggled with general climate text that could have originated from both convinced and unconvinced sources.

Our analysis of climate tweets through a six-month period in 2022 also revealed the dominant categories of contrarian claims and climate misinformation on social media relative to other information sources, such as contrarian blogs and CTT websites. While CTTs focused on policy claims and contrarian blogs focused on attacking climate science, more than half of the relevant climate tweets focused on either attacking climate actors or conspiracy theories. This result is consistent with other research finding that attacks on climate supporters are a dominant theme among tweets by climate contrarians, as are climate-themed conspiracy theories \citep{xia2021spread}. These findings are also consistent with recent scholarship on the personalisation of politics on social media \citep{bennet2012personalization, kissas2022populist} and the observation that personal attacks tend to drive engagement on platforms such as Facebook and Twitter/X \citep{rathje2021outgroup}, suggesting that the nature of misinformation---i.e., personalised versus non-personalised---is at least particularly dependent on the platform of interest. Overall, the dominance of personal attacks in climate misinformation on social media underscores the importance of better understanding the impact of climate misinformation in the form of ad hominem attacks and conspiracy theories, as well as exploring the efficacy of interventions that neutralise their negative impact.

We also identified the different types of misinformation peaks on Twitter, associated with external events (political or natural) or influencer posts (contrarian or convinced). External events coincided with a spike in the total number of climate tweets.  while influencer-associated peaks were associated with an increase in the proportion of misinformation tweets but not an increase in overall climate tweets.

There were predictable patterns in the types of arguments in response to different peak types. The clearest signal was in response to natural events, which showed a 680\% increase in category 1.7 claims making the argument that weather events weren’t linked to climate change. Political events coincided with category 4.1 claims, arguing that climate policy was harmful. 

A limitation of this study is that its scope was restricted to climate misinformation on Twitter. It is yet to be seen whether the Augmented CARDS model performs at similar levels on other data sources. Future research could focus on a wider range of information sources, such as other social media platforms, congressional testimonies, public speeches, online video transcripts, and newspaper articles. Such an analysis could also yield which misinformation categories are dominant across these different information sources. Within a single information source, cross-country analysis could also interrogate different emphases in climate misinformation across different cultures. Similarly, analysis of output from different mainstream media publishers could identify the relative proportion of climate misinformation among different outlets.

Future research may also apply supervised learning models to a broader period of time to draw out long term trends on Twitter and other platforms. For example, there has been much discussion surrounding the recent changes to Twitter's ownership \citep{HKS2023Musk} and the impact this has had on climate misinformation \citep{ISD2023COP27}. Our data does not span a large enough time period to make any concrete conclusions about these recent changes. 

Another limitation of the CARDS model is that, to date, it has been trained on English text only. Future research could apply our methodology with training sets of non-English text, to facilitate detection of climate misinformation in other languages across different countries. 

\section{Conclusion}\label{sec13}

This study has taken a step closer to the goal of automatically detecting and correcting climate misinformation in real-time. We have shown an improvement in classifying misinformation in climate tweets, with significant reductions in the "false positive problem" associated with training on contrarian actors alone. 

These findings have practical implications. Adopting our model could help Twitter/X to augment and enhance ongoing manual fact-checking procedures by offering a computer-assisted procedure for finding the tweets most likely to contain climate misinformation. This adoption could make finding and responding to climate-related misinformation more efficient and help Twitter/X enforce policies to reduce false or misleading claims on the platform. Yet environmental groups have shown that Twitter/X ranks dead last among major social media platforms in its policies and procedures for responding to climate misinformation and there is little evidence that X will improve these procedures in the near term \citep{caad2023bigtech}. Alternatively, our model could provide the basis for an API that Twitter/X users could employ to assess climate-related claims they are seeing in their feeds. Overall, the potential practical applications of our model underscores the need for continued academic work to monitor misinformation on Twitter/X and raises important questions on the data needed to hold social media platforms accountable for the spread of false claims.

However, there are still numerous hurdles to overcome before the goal of automated debunking is achieved. An effective debunking requires both explanation of the relevant facts and exposing the misleading fallacies employed by the misinformation. Contrarian climate claims can contain a range of different fallacies, so automatic detection of logical fallacies is another necessary task that, used in concert with the CARDS model, could bring us closer to the "holy grail of fact-checking" \cite{hassan2015quest}.

Regardless, this research has already provided greater understanding of climate misinformation on social media, identifying four types of misinformation spikes. Knowing the types of arguments that are likely to be posted on social media in response to external events such as climate legislation or natural events can inform interventions that seek to pre-emptively neutralize anticipated misinformation narratives.

\section*{Acknowledgements}

We gratefully acknowledge the large dataset provided by the Online Media Monitor (OMM) from the University of Hamburg and the Monash Data Futures Institute for their support.

\section*{Author Contributions Statement}

The paper's authorship contributions are as follows: John Cook and Yuan-Fang Li led the study's conceptualization and design; data collection was carried out by Frank Algra-Maschio, Cristian Rojas and Travis Coan; machine learning algorithm development and data analysis were led by Cristian Rojas and Yuan-Fang Li; Frank Algra-Maschio and YMark Andrejevic explored social correlations and interpreted results. All authors contributed to drafting the manuscript, critically reviewed the findings, and approved the final version.

\section*{Competing Interests Statement}

All authors declare no competing interests.

\bibliography{main}
\pagebreak

\begin{appendices}

\section{Data}\label{appendice:data}
\subsection{CARDS}

The CARDS dataset encompassed approximately 29,000 claims concerning climate change. Within this dataset, roughly 30\% of the claims were determined to be misinformation, using the taxonomy devised by Coan et al. The detailed taxonomy can be found in Table \ref{table:taxonomy}. A group of coders well-versed in climate-related matters labelled these claims by analyzing 87,178 paragraphs extracted from communications originating from conservative think tanks (CTTs) and central contrarian blogs \cite{coan2021computer}\cite{cook2020deconstructing}.

\begin{table}[h]
\begin{tabular}{@{}rl@{}}
\toprule
\multicolumn{1}{c}{{\color[HTML]{212529} \textbf{Code}}} & \multicolumn{1}{c}{{\color[HTML]{212529} \textbf{Claim label}}}                                               \\ \midrule
\multicolumn{1}{l}{{\color[HTML]{212529} \textbf{0}}} & {\color[HTML]{212529} No claim}                                                     \\
\multicolumn{1}{l}{{\color[HTML]{212529} \textbf{1}}}    & {\color[HTML]{212529} \textbf{Global warming is not happening}}                                               \\
{\color[HTML]{212529} 1.1}                            & {\color[HTML]{212529} Ice/permafrost/snow cover isn't melting}                      \\
{\color[HTML]{212529} 1.2}                            & {\color[HTML]{212529} We're heading into an ice age/global cooling}                 \\
{\color[HTML]{212529} 1.3}                            & {\color[HTML]{212529} Weather is cold/snowing}                                      \\
{\color[HTML]{212529} 1.4}                            & {\color[HTML]{212529} Climate hasn't warmed/changed over the last (few) decade(s)}  \\
{\color[HTML]{212529} 1.6}                            & {\color[HTML]{212529} Sea level rise is exaggerated/not accelerating}               \\
{\color[HTML]{212529} 1.7}                               & {\color[HTML]{212529} Extreme weather isn't increasing/has happened before/isn't linked to climate change}    \\
\multicolumn{1}{l}{{\color[HTML]{212529} \textbf{2}}}    & {\color[HTML]{212529} \textbf{Human greenhouse gases are not causing climate change}}                         \\
{\color[HTML]{212529} 2.1}                            & {\color[HTML]{212529} It's natural cycles/variation}                                \\
{\color[HTML]{212529} 2.3}                               & {\color[HTML]{212529} There's no evidence for greenhouse effect/carbon dioxide driving climate change}        \\
\multicolumn{1}{l}{{\color[HTML]{212529} \textbf{3}}}    & {\color[HTML]{212529} \textbf{Climate impacts/global warming is beneficial/not bad}}                          \\
{\color[HTML]{212529} 3.1}                            & {\color[HTML]{212529} Climate sensitivity is low/negative feedbacks reduce warming} \\
{\color[HTML]{212529} 3.2}                               & {\color[HTML]{212529} Species/plants/reefs aren't showing climate impacts/are benefiting from climate change} \\
{\color[HTML]{212529} 3.3}                            & {\color[HTML]{212529} CO2 is beneficial/not a pollutant}                            \\
\multicolumn{1}{l}{{\color[HTML]{212529} \textbf{4}}}    & {\color[HTML]{212529} \textbf{Climate solutions won't work}}                                                  \\
{\color[HTML]{212529} 4.1}                            & {\color[HTML]{212529} Climate policies (mitigation or adaptation) are harmful}      \\
{\color[HTML]{212529} 4.2}                            & {\color[HTML]{212529} Climate policies are ineffective/flawed}                      \\
{\color[HTML]{212529} 4.4}                            & {\color[HTML]{212529} Clean energy technology/biofuels won't work}                  \\
{\color[HTML]{212529} 4.5}                            & {\color[HTML]{212529} People need energy (e.g. from fossil fuels/nuclear)}          \\
\multicolumn{1}{l}{{\color[HTML]{212529} \textbf{5}}}    & {\color[HTML]{212529} \textbf{Climate movement/science is unreliable}}                                        \\
{\color[HTML]{212529} 5.1}                               & {\color[HTML]{212529} Climate-related science is unreliable/uncertain/unsound (data, methods \& models)}      \\
{\color[HTML]{212529} 5.2}                            & {\color[HTML]{212529} Climate movement is unreliable/alarmist/corrupt}              \\ \bottomrule
\end{tabular}
\caption{Taxonomy employed to categorize the misinformation claims within the CARDS dataset. It consists of two hierarchical levels, encompassing a total of 18 categories.}
\label{table:taxonomy}
\end{table}

\subsection{Waterloo}
The dataset compiled by the University of Waterloo consists of labelled tweets pertaining to climate change, covering the time period from April 27, 2015, to February 21, 2018. In total, 43,943 tweets were annotated. Each tweet underwent individual labelling by three reviewers, and only those tweets that received unanimous agreement from all reviewers were included \cite{Kaggle}.

\subsection{Hamburg}
The Online Media Monitor (OMM) from the University of Hamburg contributed with a dataset of  5,236,660 unlabeled tweets gathered from June 21, 2022, to December 8, 2022. The data was filtered from the platform based on keywords or phrases that included: \#climatechange, climate change, "global warming," climate crisis, or climate emergency \cite{omm2023}.

\pagebreak

\section{Anomalies: Word Analysis}\label{appendice:words}
\begin{table}[h]
    \centering
    \begin{subtable}{.55\linewidth}
     \begin{tabular}{@{}rrr@{}}
        \toprule
        \textbf{Token}    & \textbf{Log Fold Change} & \textbf{P value} \\ \midrule
        \rowcolor[HTML]{FFFFFF} 
        heatwave          & 1.474412                 & 1.13E-18         \\
        \rowcolor[HTML]{FFFFFF} 
        declare           & 1.432399                 & 7.48E-16         \\
        \rowcolor[HTML]{FFFFFF} 
        fires             & 1.006194                 & 4.70E-17         \\
        \rowcolor[HTML]{FFFFFF} 
        hot               & 0.958189                 & 2.69E-45         \\
        \rowcolor[HTML]{FFFFFF} 
        biden             & 0.906806                 & 4.52E-115        \\
        \rowcolor[HTML]{FFFFFF} 
        heat              & 0.897032                 & 8.50E-112        \\
        \rowcolor[HTML]{FFFFFF} 
        @potus            & 0.854548                 & 1.18E-50         \\
        \rowcolor[HTML]{FFFFFF} 
        summer            & 0.784148                 & 1.27E-22         \\
        \rowcolor[HTML]{FFFFFF} 
        climate emergency & 0.735294                 & 1.44E-113        \\
        \rowcolor[HTML]{FFFFFF} 
        uk                & 0.574371                 & 2.18E-17         \\ \bottomrule
        \end{tabular}
        \label{table:awordanalysis}
        \caption{\label{table:awordanalysis} }
    \end{subtable}

    \begin{subtable}{.25\linewidth}
        \begin{tabular}{@{}rrr@{}}
            \toprule
            \textbf{Token} & \textbf{Log Fold Change} & \textbf{P value} \\ \midrule
            \rowcolor[HTML]{FFFFFF} 
            ian            & 2.713324                 & 6.19E-103        \\
            \rowcolor[HTML]{FFFFFF} 
            hurricane      & 2.584126                 & 0.00E+00         \\
            \rowcolor[HTML]{FFFFFF} 
            florida        & 2.161538                 & 1.11E-88         \\
            \rowcolor[HTML]{FFFFFF} 
            @danrather     & 2.107857                 & 8.97E-23         \\
            \rowcolor[HTML]{FFFFFF} 
            storm          & 1.935989                 & 3.65E-25         \\
            \rowcolor[HTML]{FFFFFF} 
            climate change & -0.089166                & 1.43E-233        \\
            \rowcolor[HTML]{FFFFFF} 
            global warming & -0.315675                & 2.28E-88         \\
            \rowcolor[HTML]{FFFFFF} 
            us             & -0.355969                & 9.99E-28         \\
            \rowcolor[HTML]{FFFFFF} 
            global         & -0.36468                 & 1.96E-181        \\
            \rowcolor[HTML]{FFFFFF} 
            climate crisis & -0.405621                & 2.24E-45         \\ \bottomrule
        \end{tabular}
        \label{table:cwordanalysis}
        \caption{\label{table:cwordanalysis}}
    \end{subtable}
    
    \begin{subtable}{.25\linewidth}
        \begin{tabular}{@{}rrr@{}}
            \toprule
            \textbf{Token} & \textbf{Log Fold Change} & \textbf{P value} \\ \midrule
            \rowcolor[HTML]{FFFFFF} 
            egypt          & 1.827352                 & 4.13E-20         \\
            \rowcolor[HTML]{FFFFFF} 
            cop27          & 1.707518                 & 5.53E-53         \\
            \rowcolor[HTML]{FFFFFF} 
            conference     & 1.186116                 & 1.62E-13         \\
            \rowcolor[HTML]{FFFFFF} 
            nations        & 0.955196                 & 9.09E-11         \\
            \rowcolor[HTML]{FFFFFF} 
            leaders        & 0.871997                 & 8.68E-11         \\
            \rowcolor[HTML]{FFFFFF} 
            countries      & 0.720918                 & 1.46E-26         \\
            \rowcolor[HTML]{FFFFFF} 
            climate crisis & -0.1728                  & 2.26E-11         \\
            \rowcolor[HTML]{FFFFFF} 
            people         & -0.270252                & 3.04E-33         \\
            \rowcolor[HTML]{FFFFFF} 
            climate change & -0.29039                 & 0.00E+00         \\ \bottomrule
        \end{tabular}
        \label{table:dwordanalysis}
        \caption{\label{table:dwordanalysis} }
    \end{subtable}
    \caption{ Top 10 words ranked by their Log Fold change during the evaluated period, in comparison to the overall timeline.  (\subref{table:awordanalysis}). July 18 and July 21, 2022. (\subref{table:cwordanalysis}). September 29, 2022. (\subref{table:dwordanalysis}). November 7 and November 8, 2022.}
    \label{table:wordanalysis}
\end{table}




\end{appendices}


\end{document}